\documentclass[runningheads]{llncs}
\usepackage[T1]{fontenc}
\usepackage{graphicx}
\usepackage{color}
\usepackage{hyperref}

\usepackage{amsfonts}
\usepackage{amsmath}

\usepackage{amsmath,amsfonts,bm}

\def\eqref#1{equation~\ref{#1}}

\def\1{\bm{1}}

\def\vx{{\bm{x}}}

\DeclareMathAlphabet{\mathsfit}{\encodingdefault}{\sfdefault}{m}{sl}
\SetMathAlphabet{\mathsfit}{bold}{\encodingdefault}{\sfdefault}{bx}{n}

\usepackage{booktabs}   
\usepackage{multirow}

\begin{document}
\title{Learning Temporally Equivariance for Degenerative Disease Progression in OCT by Predicting Future Representations}


%
%
\author{Taha Emre\and
Arunava Chakravarty\and
Dmitrii Lachinov\and
Antoine Rivail\and\\
Ursula Schmidt-Erfurth\and
Hrvoje Bogunovi\'c}
%

\authorrunning{Emre et al.}
%
\institute{Department of Ophthalmology and Optometry, Medical University of Vienna, Austria\\
\email{\{taha.emre,hrvoje.bogunovic\}@meduniwien.ac.at}}
\maketitle              
\begin{abstract}

Contrastive pretraining provides robust representations by ensuring their invariance to different image transformations while simultaneously preventing representational collapse. Equivariant contrastive learning, on the other hand, provides representations sensitive to specific image transformations while remaining invariant to others. By introducing equivariance to time-induced transformations, such as disease-related anatomical changes in longitudinal imaging, the model can effectively capture such changes in the representation space. In this work, we propose a Time-equivariant Contrastive Learning (TC) method. First, an encoder embeds two unlabeled scans from different time points of the same patient into the representation space. Next, a temporal equivariance module is trained to predict the representation of a later visit based on the representation from one of the previous visits and the corresponding time interval with a novel regularization loss term while preserving the invariance property to irrelevant image transformations. On a large longitudinal dataset, our model clearly outperforms existing equivariant contrastive methods in predicting progression from intermediate age-related macular degeneration (AMD) to advanced wet-AMD within a specified time-window.

\end{abstract}
\section{Introduction}

The advent of contrastive self-supervised learning (SSL) methods showed that it is possible to learn informative and discriminative image representations by learning invariances to image transformations that do not alter their semantics. The invariance is achieved by increasing the representational similarity between the original image and its transformed versions in the representation space, while pushing all other pairs apart as negatives using a Siamese network~\cite{azizi2021big}. 
However, sensitivity (\textit{equivariance}) to certain transformations may be crucial for specific downstream tasks, such as rotation degree information in histopathology slides~\cite{veeling2018rotation}. Introducing equivariance to time-induced transformations, such as anatomical changes due to degenerative disease progression in longitudinal medical images could be pivotal in downstream tasks such as predicting the future risk of conversion of a patient from an early to a late-stage. 

Retinal degenerative diseases  such as Age-related Macular Degeneration (AMD) are frequently identified through non-invasive optical coherence tomography (OCT) scans, 
made up of sequential 2D cross-sectional slices (B-scans). As a leading cause of vision loss and blindness~\cite{blii}, AMD leads to irreversible tissue damage, implying that the severity of the disease can be represented as a monotonically non-decreasing function over time. However, the speed of disease progression and the associated anatomical changes vary widely between patients. Also, the longitudinal nature of the AMD datasets makes them costly to label by clinicians. Consequently, a Time-Equivariant SSL stage could be crucial in identifying the patients at a higher risk of progression to an advanced disease stage from a single visit to provide timely treatment and personalized management.

Existing \textit{Equivariant SSL} methods aim to learn such sensitivity, unlike contrastive methods that seek total invariance. Indeed, invariance to certain transformations is not always desirable~\cite{xiao2020should}. This sensitivity can be achieved with pretext SSL, by learning to recover the parameter of the target transformation from the representations. Image rotation prediction~\cite{gidaris2018unsupervised}, video clip ordering~\cite{xu2019self} and distinguishing close and far clips~\cite{jayaraman2016slow} for temporal equivariance can be given as examples. In medical imaging, the time difference~\cite{rivail2019modeling} or ordering two visits (scans) of a patient~\cite{kim2023learning} is commonly used as a pretraining step for temporal tasks such as disease forecasting. In~\cite{ZHAO2021longssl} they decoupled the effect of aging from disease-related changes in brain MRI scans by carefully curating a temporal dataset consisting of old subjects without any underlying diseases. Then, they learned an aging trajectory in the representation space shared by all patients. Although these methods learn sensitivity to the transformation, they lack the desired invariance to image perturbations.

To address these issues, \textit{Equivariant contrastive methods} have been explored to learn equivariance for specific transformations while ensuring invariance for the others. Unlike methods with specific architecture~\cite{grequi} for achieving equivariance, equivariant contrastive learning methods are mostly architecture agnostic. Dangovski et al. \cite{dangovski2022equivariant} combined transformation parameter prediction with a contrastive loss for the transformation invariance, requiring four additional encoders for the rotation prediction task~\cite{gidaris2018unsupervised}, increasing the computational overhead substantially (\textit{ESSL}). Similarly, in~\cite{jenni2021time}, they proposed an additional temporal pretext task, such as frame ordering, to the contrastive task for videos. Emre et al.~\cite{emre2022tinc} introduced \textit{TINC} for OCT scans where they exploited the time difference between visits by applying a similarity insensitivity increasing with the difference. In \cite{lee2021improving}, \textit{AugSelf} is proposed to enforce equivariance on the representation space by predicting transformation parameters between input pairs, then projected the learned representations into an invariant space for the contrastive task. However, transformation parameter recovery does not guarantee explicit equivariance, because it's possible for transformation parameters to be confined to a single dimension of the representation, leaving the rest completely invariant. In addition, parameter prediction models do not learn any explicit function for directly transforming the representations. To address these limitations, Devillers et al.~\cite{devillers2023equimod} proposed utilizing two different projections of the representation space; one for the invariance, and the second for the equivariance with a trainable module allowing direct manipulation of the projections with the image transformation parameter (\textit{EquiMod}). Since invariance is a trivial solution for the equivariance where the equivariant representation transformation collapses to identity, they introduced an additional contrastive loss for the equivariance branch at the cost of an increased computational overhead. These methods aim to achieve equivariance using transformation parameter recovery, additional contrastive tasks, projection spaces, or pre-defined image transformation (e.g. rotation matrix)~\cite{garrido2023sie}. 

We propose to learn time-sensitive/equivariant representations directly on the \textit{representation space} and learn to propagate representations in time, using longitudinal medical datasets in a contrastive SSL setting. Given that degenerative diseases either progress or stagnate over time, we have modeled disease progression in the representation space as an additive transformation parameterized by the time difference of a patient's visits, enabling direct temporal transformation of representations. We adapted a regularization loss on the calculated additive term to prevent the network from ignoring the time difference and producing identity transformation. Finally, we evaluated the learned representations for AMD progression prediction across multiple time-windows from a single visit scan. In summary, \textbf{our contributions} are (a) constructing contrastive pairs from different visits of a patient, (b) training a novel equivariance module to propagate representation in the future without accessing future visit scan, (c) introducing a novel regularization loss to prevent invariance to time.

\section{Methodology}
\begin{figure}[hbt]
\begin{center}
\includegraphics[width=0.95\linewidth]{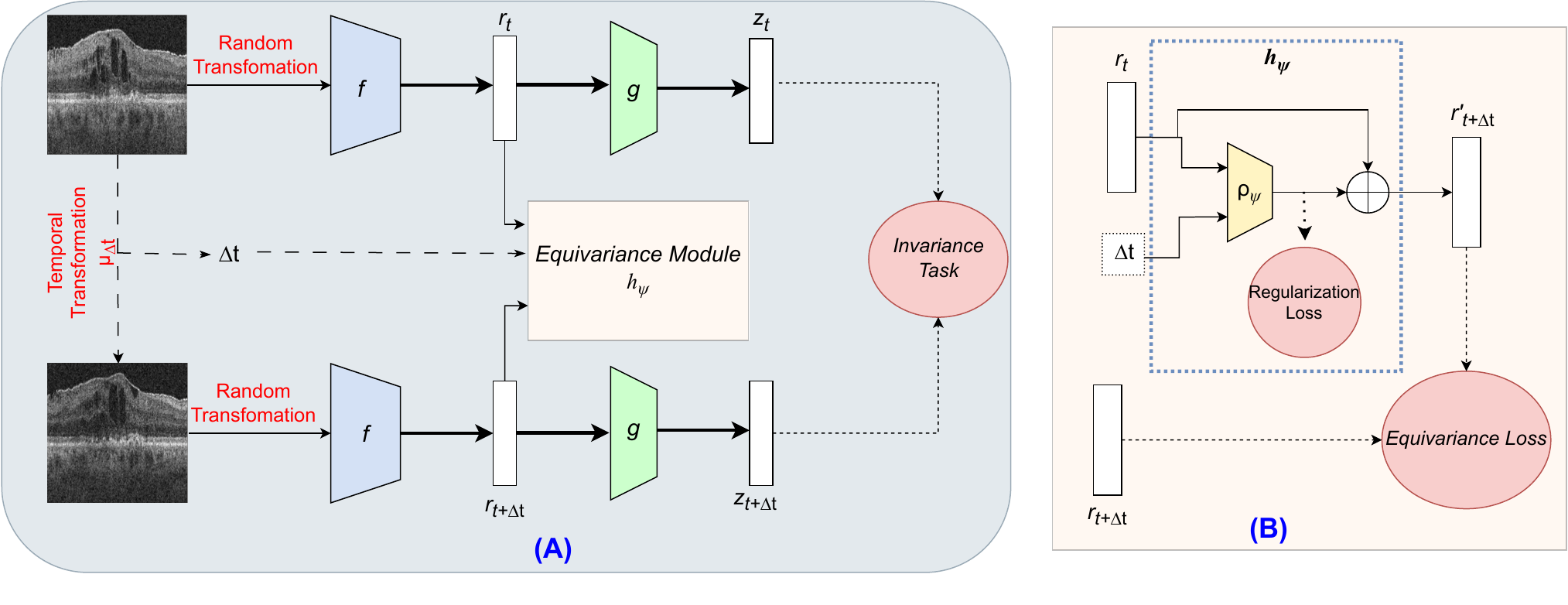}
\end{center}
\caption{An overview of \textbf{TC} architecture. \textbf{(A)} Representations of two scans acquired at two time points and the time difference are fed into $h_\psi$ for equivariance, then projected for the invariance loss. \textbf{(B)} Equivariance module with an additive displacement map.}
\label{fig:mainfig}
\end{figure}

Our approach learns temporal equivariance in representation space, followed by a projection to invariant space within a contrastive SSL framework (Fig.~\ref{fig:mainfig}). We propose a novel loss term specifically tailored for the equivariance (Eq.~\ref{eq:finak}), utilizing a longitudinal retinal OCT dataset (Sec.~\ref{sec:datasets}) collected across multiple patient visits. First, we define the term time equivariant representation. Then, we detail our approach to learn the equivariance. 
\subsubsection{Time-Equivariant Representations}

The disease progression in a patient is routinely monitored using a series of medical scans $\vx \in \mathbb{I} \subseteq \mathbb{R}^{H\times W}$, and $\vx_t$ denotes the scan acquired at a time $t$ sampled from  $\mathbb{T} = \lbrace t \in \mathbb {N}_0; t \leq b \rbrace$,  $b$ being the last visit date. Let $\mu : \mathbb{I}\times\mathbb{T}\rightarrow\mathbb{I}$ represent a transformation in the image space for all possible time differences, which captures the complex anatomical changes between $\vx_t$ and $\vx_{t+\Delta t}$ due to the disease progression (and/or normal aging), where $t+\Delta t \in \mathbb{T}$. Although the actual transformation $\mu$ is difficult to model in the image space, the result of the transformation is available as $\vx_{t+\Delta t}$ for some specific $\Delta t$, from available visits of the patient, such that $\vx_{t+\Delta t} := \mu(\vx_t,\Delta t) $. Let $f_\theta:\mathbb{I}\rightarrow\mathbb{R}^{D}$ with learnable parameters $\theta$ be a deep learning based encoder, which can embed $\vx_t$ as a representation $r_t := f_\theta(\vx_t) $ and $\vx_{t+\Delta t}$ as $r_{t+\Delta t} := f_\theta(\vx_{t+\Delta t}) $. As $\mu(\cdot , \Delta t)$ transforms $\vx_{t}$ to $\vx_{t+\Delta t}$ in the image space, a corresponding transformation $h(\cdot , \Delta t)$  operates in the representation space such that $r_{t+\Delta t} = h(r_t,\Delta t) $. If $f_\theta$ is equivariant in time, the \textbf{equivariant relation} is defined as:

\begin{equation}
    \label{eq:equiv}
   \exists h : \forall \Delta t \in \mathbb{T},  \quad f_\theta(\mu(\vx_t,\Delta t)) =  f_\theta(\vx_{t+\Delta t})  \approx  h(f_\theta(\vx_t), \Delta t) 
\end{equation}

An undesired trivial solution for Eq. \ref{eq:equiv} would be \textit{time-invariance} of $f_\theta$ such that $f_\theta(\vx_t) = f_\theta(\vx_{t+\Delta t})$ with disease progression being ignored, in turn making $h(\cdot , \Delta t)$ an identity mapping. This implies that $f_{\theta}$ learns only patient-specific features staying constant with respect to time while ignoring changes due to the progression. The transformation $h(\cdot , \Delta t)$ should be able to capture the future changes in the scans specific to each patient, and we propose to approximate it with a deep neural network that can model complex changes.

\subsubsection{Time-Equivariant Contrastive Learning (TC)} \label{Section:proj} Our model extends the existing contrastive methods with the introduction of a temporal equivariance property. The input pair to the Siamese network is created from two different visits of a patient and both are transformed with contrastive augmentations~(Fig.\ref{fig:mainfig}.A). Recent equivariant contrastive methods with a learnable transformation~\cite{devillers2023equimod,garrido2023sie} on features, relied on two projection spaces, one for equivariance and the other for invariance, preventing them from directly steering the representations. Instead, we propose to enforce equivariance directly on the representation space, and then to learn the invariance on a \textit{projection} space~\cite{jing2022understanding} (Fig.\ref{fig:mainfig}.A). This allows for the preservation of the time-sensitive information in the representation space while enforcing invariance in the projection space. A logical order since invariance is a trivial solution of equivariance. It also simplifies the network architecture and computational overhead, and facilitates transferring the encoder weights as a backbone.

We introduce an equivariance module $h_\psi$ containing the learned transformation (\textbf{predictor}) with learnable parameters $\psi$, which takes the concatenation of representation $r_t$ and time difference $\Delta t$, normalized between 0-1, as input, and transforms $r_t$ to $r^\prime_{t+\Delta t}$ to satisfy the equivariance property defined in Eq. \ref{eq:equiv}. The predictor enables the model to generate future visits' representations by forwarding available scans in time to be used in predictive tasks. The equivariance loss~\cite{devillers2023equimod,garrido2023sie} for $h_\psi$ is defined as:
\begin{equation}
    \label{eq:equiloss}
     \ell_{\text{equiv}} =\|   f_\theta(\vx_{t+\Delta t}) -  h_\psi(f_\theta(\vx_{t}),\Delta t) \|_2^2 = \|   r_{t+\Delta t} -  h_\psi(r_{t},\Delta t) \|_2^2
\end{equation}

As it is highlighted in \cite{devillers2023equimod,garrido2023sie}, the contrastive loss imposes invariance to the image transformations including time-related changes. The learned invariance results in a trivial solution for the equivariance loss where $h_\psi$ collapses to an identity mapping,  ignoring $\Delta t$. The aforementioned methods rely on another set of computationally heavy contrastive task on $r^\prime_{t+\Delta t}$ to prevent collapse. For an MLP $h_\psi$, the collapse manifests as $\frac{\partial h_{\psi}}{\partial t}$ becoming 0. To avoid an additional contrastive task or a regularization term on the gradients to prevent $h_\psi$ from collapsing, we reparametrized $h_{\psi}$ as an additive \textbf{displacement map} (DM):

\begin{equation}
    \label{eq:displ}
    r^\prime_{t+\Delta t} = h_\psi(r_t,\Delta t) = r_t \oplus  \rho_\psi(r_t,\Delta t) 
\end{equation}
Now, the equivariant transformation $h_\psi$ has two parts: an MLP $\rho_\psi$ for predicting DM as an additive term and \textit{direct sum} of the DM with the representation (Fig.~\ref{fig:mainfig}.B). In Eq.~\ref{eq:displ}, the collapse condition is much easier to regularize since it is the zero norm for the predicted DM. The ideal regularization should prevent DM norm becoming zero, while not encouraging high norm values. The representation space should respond to the time change smoothly without predicting large DMs easily. We adapted a regularization loss term for preventing collapse from pair-wise loss of RankNet~\cite{burges2005rank}. The original loss function aims to increase the probability of ranking two embeddings correctly. In our case, the ranking is always one-sided, because the norm cannot take negative values. Thus, the final DM regularization term becomes:

\begin{equation}
    \label{eq:reg}
    \ell_{\text{reg}} = \log \left(1 + \exp(-||  \rho_\psi(r_t,\Delta t) ||_2) \right)
\end{equation}
The $\ell_{\text{reg}}$ penalizes null displacements without encouraging very high norms. 

For the invariant contrastive loss, we used VICReg~\cite{bardes2022vicreg}. It is calculated using projections $z_t$ and $z_{t+\Delta t}$, obtained by projecting representations $r_t$ and $r_{t+\Delta t}$ with an MLP projector $g_{\gamma}$. As a \textit{non-contrastive} method, it does not require very large batch sizes. It has 3 loss terms; an $\ell_{2}$ term for increasing similarity within pair, a covariance term to prevent \textit{dimensional collapse}~\cite{zbontar2021barlow,jing2022understanding} and a variance term for implicitly pushing the negatives apart which is calculated along a batch of projections, where each pair could have different $\Delta t$ values.

Thus, the final training loss with a contrastive loss $\ell_{\mathrm{contr}}$ is:
\begin{equation}
    \label{eq:finak}
   \mathcal{L}_{\mathrm{TC}} = \ell_{\mathrm{contr}} + \beta \cdot ( \ell_{\text{equiv}} + \upsilon \cdot \ell_{\text{reg}})
\end{equation}

\section{Experiments and Results}
We tested \textbf{TC} against three main equivariant contrastive methods: ESSL~\cite{dangovski2022equivariant}, AugSelf~\cite{lee2021improving} and EquiMod~\cite{devillers2023equimod}. ImageNet pretrained ResNet50, VICReg~\cite{bardes2022vicreg} and TINC~\cite{emre2022tinc} were used as baselines. During pretraining each batch contains only a single image pair from a patient to avoid pushing apart the intra-patient scans with the contrastive loss. 
Clinically, only future representations matter for conversion risk assessment. Therefore, the equivariant models were pre-trained only with image pairs with a positive  $\Delta t$.
\subsection{Experimental Setup}
\subsubsection{Dataset}
\label{sec:datasets}
\begin{figure}[t]
\begin{center}
\includegraphics[width=0.85\linewidth]{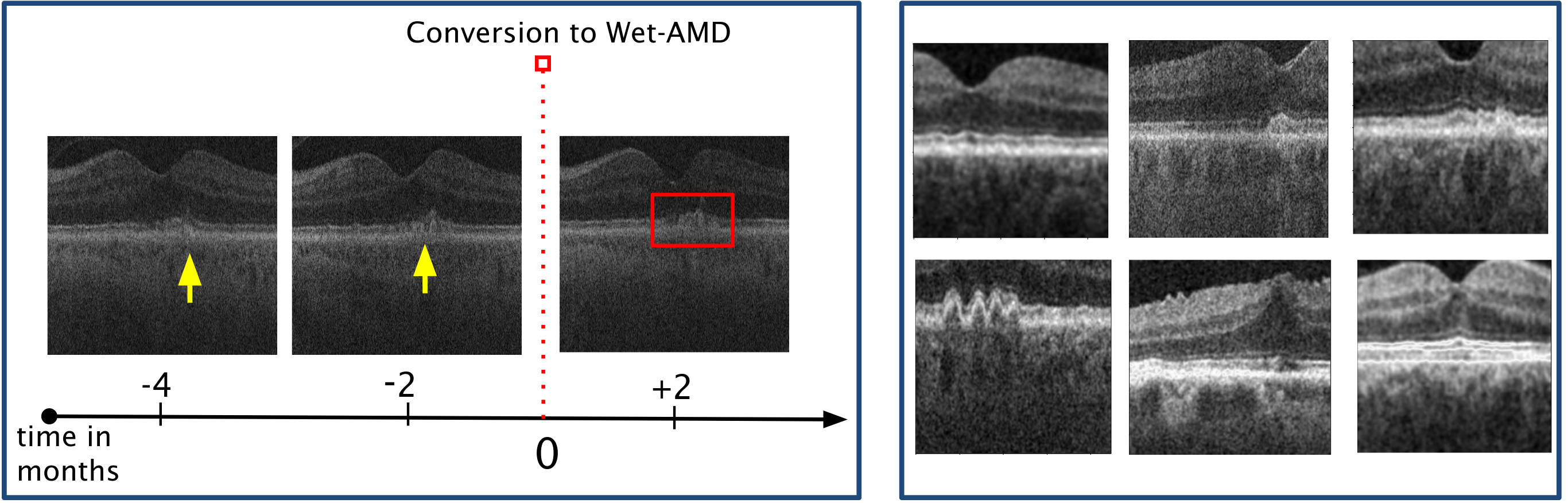}
\end{center}
\caption{\textbf{Left}: OCT scans with conversion to wet-AMD over time. \textbf{Right}: Contrastive augmentations examples.}
\label{fig:datasets}
\end{figure}
The SSL stage and the downstream evaluation were conducted on the HARBOR dataset (Fig.~\ref{fig:datasets} left), comprising OCT scans from 1,096 patients' \textit{fellow-eyes}\footnote{The other eye that is not part of the clinical trial} monitored for wet-AMD onset, with monthly follow-ups over 24 months. Intermediate stage visits (463 eyes - 10,108 scans) are used for predicting the onset of late, wet-AMD stage within 6 and 12 months~\cite{yan2020deep,yim2020predicting,russakoff2019deep} (117 converter eyes) as a forecasting task. The dataset was split with the eye level stratification into 4-fold cross-validation ($80\%$) for hyper-parameter optimization and a hold-out test set ($20\%$) for reporting the performance. The remaining 540 eyes (12,506 scans) are only used in contrastive pretraining step. We used the fovea centered \textbf{2D B-scan}, which is known to be most representative of the AMD state~\cite{fove}, resized to $224 \times 224$.

We followed the work of~\cite{emre2022tinc,met} for contrastive image transformations specific to B-scans (Fig.~\ref{fig:datasets} right). During pretraining, for each input pair, $\Delta t$ is \textbf{randomly}  sampled from an interval of 1 to 12 months and normalized between $0-1$. The rationale behind the interval is that the set of all pair permutations for all possible time differences is very large (276 per patient) for a plausible training time. By limiting $\Delta t$, training $h_\psi$ becomes much more convenient, because it only needs to produce DMs for a constrained but practically meaningful interval. 

\subsubsection{Implementation details}
A ResNet50 with a representation dimension of 2048 is used as the encoder $f_\theta$. Following the common practice of projecting representations to higher dimensions~\cite{bardes2022vicreg,zbontar2021barlow}, the projector $g_{\gamma}$ is a three-layer MLP with hidden dimensions of 4096. DM predictor $\rho_\psi$ of TC is implemented as a two-layer MLP with an input size of ($2048 + 1$) for $[r_{t}, \Delta t]$, and it outputs a DM with the same dimension as the representations. In EquiMod, ESSL and AugSelf, we followed the original works for the predictor architecture. 

All models were pretrained for 300 epochs with the AdamW optimizer with a learning rate of 5E-4 and a batch size of 128. A cosine scheduler with a warm-up was used for decaying learning rates. The same weight decay of 1E-6 was applied to all pretraining setups. $\beta$ and $\upsilon$ of $\mathcal{L}_{\mathrm{TC}}$ were set to 1 and 0.5, respectively.
\subsection{Results}
\setlength{\tabcolsep}{3pt}
\begin{table}[!t]
\centering
\caption{Linear evaluation results for wet-AMD conversion prediction within two time-windows: 6 and 12 months}\label{Tbl:Results}
\resizebox{1.0\textwidth}{!}
{
\begin{tabular}{cllllll}
\multirow{2}{*}{\textbf{Model}}  & \multicolumn{3}{c}{\textbf{6-months}} & \multicolumn{3}{c}{\textbf{12-months}} \\
\cmidrule(lr){2-4} \cmidrule(lr){5-7}
&  \multicolumn{1}{c}{AUROC $\uparrow$} & \multicolumn{1}{c}{PRAUC $\uparrow$} & \multicolumn{1}{c}{BAcc $\uparrow$} & \multicolumn{1}{c}{AUROC $\uparrow$} & \multicolumn{1}{c}{PRAUC $\uparrow$} & \multicolumn{1}{c}{BAcc $\uparrow$} \\ \midrule
ImageNet&$0.714\pm0.018$&$0.106\pm 0.007$ &$0.648\pm0.016$&$0.692\pm0.014$  & $0.167\pm0.009$ & $0.616\pm0.045$\\
VICReg\cite{bardes2022vicreg}&$0.700\pm0.029$&$0.137\pm 0.031$ &$0.650\pm0.021$&$0.671\pm0.033$  & $0.158\pm0.019$ & $0.622\pm0.023$\\
TINC\cite{emre2022tinc}&$0.729\pm0.017$&$0.151\pm 0.019$ &$0.652\pm0.007$&$0.688\pm0.060$  & $0.210\pm0.026$ & $0.629\pm0.055$\\\hline
ESSL\cite{dangovski2022equivariant}&$0.718\pm0.033$&$ 0.157\pm 0.034$  &$0.649\pm0.028$&$0.699\pm0.041$&$0.187\pm0.039$&$0.637\pm0.032$ \\
AugSelf\cite{lee2021improving} &$0.752\pm0.031$& $\mathbf{0.190\pm 0.027}$  &$0.670\pm0.029$&$0.735\pm0.049$  &$0.245\pm0.040$ &$0.651\pm0.057$\\
EquiMod\cite{devillers2023equimod}&$0.734\pm0.035$&$0.153\pm0.038$  &$0.652\pm0.048$& $0.718\pm0.059$& $ 0.226\pm 0.055$ & $0.657\pm0.031$ \\
\hline
TC w/o \textit{DM} &$0.761\pm0.032$&$0.163\pm0.027$&$0.658\pm0.034$&$0.748\pm0.033$&$0.220\pm0.026$&$0.653\pm0.031$\\\
TC w/o $\ell_{\text{reg}}$&$0.732\pm0.021$&$0.137\pm0.024$&$0.661\pm0.009$&$0.721\pm0.035$&$0.195\pm0.025$&$0.655\pm0.028$\\ 
TC& $\mathbf{0.769\pm0.027}$& $0.185\pm0.032$& $\mathbf{0.677\pm0.026}$& $\mathbf{0.752\pm0.034}$ & $\mathbf{0.265\pm0.043}$& $\mathbf{0.671\pm0.032}$\\\hline
TC$_{syn}$& $\underline{0.784\pm0.022}$& $\underline{0.188\pm0.036}$& $\underline{0.683\pm0.024}$& $\underline{0.775\pm0.030}$ & $\underline{0.304\pm0.052}$& $\underline{0.695\pm0.023}$\\
\bottomrule
\end{tabular}}
\end{table}
\subsubsection{Downstream prediction performance}
After pretraining, we evaluated the extracted representations with a linear classifier on the downstream tasks (Sec.~\ref{sec:datasets}), as it is common practice in contrastive SSL~\cite{chenempirical}. A scan from a single time-point is used for future conversion prediction. The linear classifier is trained for 50 epochs with Adam optimizer and a learning rate of 1E-4. We reported the results (Table~\ref{Tbl:Results}) with AUROC, PRAUC and Balanced Accuracy (BAcc) due to a large class imbalance (6-month 1:20, 12-month 1:10). Accordingly, the random prediction PRAUC for 6-month is 0.05, and for 12-month is 0.11.

Our TC model surpassed all the other ones in AUROC and BAcc in both tasks. In terms of 12-month PRAUC, TC outperformed all while achieving competitive 6-month PRAUC performance. Interestingly, AugSelf achieved better results against the comparison methods, which can be attributed to its enforcement of the equivariance in the representation space rather than the projection space. A decline in 12-month prediction accuracy was noted for all the models, likely due to the extended time frame and the fact that the predictive biomarkers were more subtle. Finally, the performance gains of equivariant contrastive methods over the others (Table~\ref{Tbl:Results} line 1-3) underscore the importance of including temporal equivariance in disease progression prediction.

\subsubsection{Evaluation of the equivariance module}\label{Sec:equrep}
Each scan representation is synthetically propagated in time by 6 months using equivariance module $h_\psi$ without accessing the future scans. Then they were combined with the initial representations by averaging. Their predictive capability was assessed using the same linear evaluation, with the improved performances in Table~\ref{Tbl:Results} - TC$_{syn}$ demonstrating the equivariance module's ability to discern significant disease progression from a single scan. Notably, this performance increase was more distinct in the 12-month task, where the biomarkers are less distinguishable due to the longer conversion time-frame. Similar improvement is not possible with the other equivariant contrastive methods since they cannot manipulate the representation space.
%

\subsubsection{Ablation study} \label{sec:abl}
We investigated the impact of the explicit DM prediction (Table \ref{Tbl:Results} - TC w/o \textit{DM}) and the regularization loss term (Table \ref{Tbl:Results} - TC w/o $\ell_{\text{reg}}$) components of TC by removing each at the time. In both cases, the downstream task performance degraded considerably underscoring the importance of both components. When compared against the non-equivariant methods, the ablated models still perform better highlighting the importance of time component. When TC w/o \textit{DM} compared against TC w/o $\ell_{\text{reg}}$, it performs considerably better. This can be attributed to TC w/o \textit{DM} directly predicting the future representation without relying on an unregularized DM. This highlights the significance of $\ell_{\text{reg}}$ to predict meaningful DMs. 

\section{Conclusion}
Predicting late-stage disease onset is challenging due to the varying progression speeds among patients and subtle prognostic biomarkers. In this study, we proposed exploiting temporal information in unlabeled longitudinal OCT datasets. We introduced Time-equivariant Contrastive (TC) SSL with a learnable equivariance module for directly propagating image representations in time. The module retains characteristics of disease progression, such as irreversible tissue loss or degeneration, over time. A novel regularization loss was introduced to avoid the trivial solution for the equivariance induced by the contrastive loss with minimal computational overhead, unlike other equivariant models. We tested TC for disease progression on a longitudinal dataset consisting of eye scans with different stages of AMD. The TC pretraining consistently outperformed other equivariant contrastive methods as well as non-equivariant ones. Results highlight the importance of temporal sensitivity in representations for accurately assessing a patient's risk of conversion to a late disease stage. Furthermore, we found that integrating the synthetically propagated representations with the training data enhanced the conversion prediction performance, illustrating not only that TC captured relevant longitudinal information but also provided a functional prediction module, demonstrating its ability to generate a proper time-equivariant representation space.
\begin{credits}
\subsubsection{\ackname} The work has been partially funded by FWF Austrian Science Fund (FG 9-N), and a Wellcome Trust Collaborative Award (PINNACLE Ref. 210572/Z/18/Z).
\subsubsection{\discintname}The authors have no competing interests to declare that are relevant to the content of this article.
\end{credits}
%
%
%
 \bibliographystyle{splncs04}
 \bibliography{eq}

\begin{thebibliography}{10}
\providecommand{\url}[1]{\texttt{#1}}
\providecommand{\urlprefix}{URL }
\providecommand{\doi}[1]{https://doi.org/#1}

\bibitem{azizi2021big}
Azizi, S., Mustafa, B., Ryan, F., Beaver, Z., Freyberg, J., Deaton, J., Loh, A., Karthikesalingam, A., Kornblith, S., Chen, T., et~al.: Big self-supervised models advance medical image classification. In: Proceedings of the IEEE/CVF international conference on computer vision. pp. 3478--3488 (2021)

\bibitem{bardes2022vicreg}
Bardes, A., Ponce, J., LeCun, Y.: Vicreg: Variance-invariance-covariance regularization for self-supervised learning. In: International Conference on Learning Representations (2022)

\bibitem{blii}
Bressler, N.M.: {Age-Related Macular Degeneration Is the Leading Cause of Blindness . . .} JAMA  \textbf{291}(15),  1900--1901 (04 2004). \doi{10.1001/jama.291.15.1900}, \url{https://doi.org/10.1001/jama.291.15.1900}

\bibitem{burges2005rank}
Burges, C., Shaked, T., Renshaw, E., Lazier, A., Deeds, M., Hamilton, N., Hullender, G.: Learning to rank using gradient descent. In: International Conference on Machine Learning. pp. 89-- 96 (2005)

\bibitem{chenempirical}
Chen, X., Xie, S., He, K.: An empirical study of training self-supervised vision transformers. In: Proceedings of the IEEE/CVF international conference on computer vision. pp. 9620--9629 (2021)

\bibitem{grequi}
Cohen, T.S., Welling, M.: Group equivariant convolutional networks. In: International Conference on Machine Learning. PMLR, JMLR.org (2016)

\bibitem{dangovski2022equivariant}
Dangovski, R., Jing, L., Loh, C., Han, S., Srivastava, A., Cheung, B., Agrawal, P., Soljacic, M.: Equivariant self-supervised learning: Encouraging equivariance in representations. In: International Conference on Learning Representations (2022), \url{https://openreview.net/forum?id=gKLAAfiytI}

\bibitem{devillers2023equimod}
Devillers, A., Lefort, M.: Equimod: An equivariance module to improve visual instance discrimination. In: International Conference on Learning Representations (2023), \url{https://openreview.net/forum?id=eDLwjKmtYFt}

\bibitem{emre2022tinc}
Emre, T., Chakravarty, A., Rivail, A., Riedl, S., Schmidt-Erfurth, U., Bogunovi{\'c}, H.: Tinc: Temporally informed non-contrastive learning for disease progression modeling in retinal oct volumes. In: International Conference on Medical Image Computing and Computer-Assisted Intervention. pp. 625--634. Springer (2022)

\bibitem{garrido2023sie}
Garrido, Q., Najman, L., Lecun, Y.: Self-supervised learning of split invariant equivariant representations. In: International Conference on Machine Learning. PMLR (2023)

\bibitem{gidaris2018unsupervised}
Gidaris, S., Singh, P., Komodakis, N.: Unsupervised representation learning by predicting image rotations. In: International Conference on Learning Representations (2018)

\bibitem{met}
Holland, R., et~al.: Metadata-enhanced contrastive learning from retinal optical coherence tomography images. CoRR  \textbf{abs/2208.02529} (2022)

\bibitem{jayaraman2016slow}
Jayaraman, D., Grauman, K.: Slow and steady feature analysis: higher order temporal coherence in video. In: Proceedings of the IEEE/CVF Conference on Computer Vision and Pattern Recognition. pp. 3852--3861 (2016)

\bibitem{jenni2021time}
Jenni, S., Jin, H.: Time-equivariant contrastive video representation learning. In: Proceedings of the IEEE/CVF International Conference on Computer Vision. pp. 9970--9980 (2021)

\bibitem{jing2022understanding}
Jing, L., Vincent, P., LeCun, Y., Tian, Y.: Understanding dimensional collapse in contrastive self-supervised learning. In: International Conference on Learning Representations (2022), \url{https://openreview.net/forum?id=YevsQ05DEN7}

\bibitem{kim2023learning}
Kim, H., Sabuncu, M.R.: Learning to compare longitudinal images. In: Medical Imaging with Deep Learning (2023)

\bibitem{lee2021improving}
Lee, H., Lee, K., Lee, K., Lee, H., Shin, J.: Improving transferability of representations via augmentation-aware self-supervision. In: Beygelzimer, A., Dauphin, Y., Liang, P., Vaughan, J.W. (eds.) Advances in Neural Information Processing Systems (2021), \url{https://openreview.net/forum?id=U34rQjnImpM}

\bibitem{fove}
Lin, A.C., Lee, C.S., Blazes, M., Lee, A.Y., Gorin, M.B.: Assessing the clinical utility of expanded macular octs using machine learning. Translational vision science \& technology  \textbf{10}(6),  32--32 (2021)

\bibitem{rivail2019modeling}
Rivail, A., Schmidt-Erfurth, U., Vogl, W.D., Waldstein, S.M., Riedl, S., Grechenig, C., Wu, Z., Bogunovic, H.: Modeling disease progression in retinal octs with longitudinal self-supervised learning. In: International Workshop on PRedictive Intelligence In MEdicine. pp. 44--52. Springer (2019)

\bibitem{russakoff2019deep}
Russakoff, D.B., Lamin, A., Oakley, J.D., Dubis, A.M., Sivaprasad, S.: Deep learning for prediction of amd progression: a pilot study. Investigative ophthalmology \& visual science  \textbf{60}(2),  712--722 (2019)

\bibitem{veeling2018rotation}
Veeling, B.S., Linmans, J., Winkens, J., Cohen, T., Welling, M.: Rotation equivariant cnns for digital pathology. In: Medical Image Computing and Computer Assisted Intervention--MICCAI 2018: 21st International Conference, Granada, Spain, September 16-20, 2018, Proceedings, Part II 11. pp. 210--218. Springer (2018)

\bibitem{xiao2020should}
Xiao, T., Wang, X., Efros, A.A., Darrell, T.: What should not be contrastive in contrastive learning. In: International Conference on Learning Representations (2020)

\bibitem{xu2019self}
Xu, D., Xiao, J., Zhao, Z., Shao, J., Xie, D., Zhuang, Y.: Self-supervised spatiotemporal learning via video clip order prediction. In: Proceedings of the IEEE/CVF Conference on Computer Vision and Pattern Recognition. pp. 10334--10343 (2019)

\bibitem{yan2020deep}
Yan, Q., Weeks, D.E., Xin, H., Swaroop, A., Chew, E.Y., Huang, H., Ding, Y., Chen, W.: Deep-learning-based prediction of late age-related macular degeneration progression. Nature machine intelligence  \textbf{2}(2),  141--150 (2020)

\bibitem{yim2020predicting}
Yim, J., Chopra, R., Spitz, T., Winkens, J., Obika, A., Kelly, C., Askham, H., Lukic, M., Huemer, J., Fasler, K., et~al.: Predicting conversion to wet age-related macular degeneration using deep learning. Nature Medicine  \textbf{26}(6),  892--899 (2020)

\bibitem{zbontar2021barlow}
Zbontar, J., Jing, L., Misra, I., LeCun, Y., Deny, S.: Barlow twins: Self-supervised learning via redundancy reduction. In: International Conference on Machine Learning. pp. 12310--12320. PMLR (2021)

\bibitem{ZHAO2021longssl}
Zhao, Q., Liu, Z., Adeli, E., Pohl, K.M.: Longitudinal self-supervised learning. Medical Image Analysis  \textbf{71},  102051 (2021). \doi{https://doi.org/10.1016/j.media.2021.102051}, \url{https://www.sciencedirect.com/science/article/pii/S1361841521000979}

\end{thebibliography}
%

\appendix
\section{Appendix}\label{Sec:app}

\begin{table}[b]
\caption{Contrastive Augmentations and OCT Details}
\label{Tbl:augs}
\begin{center}
\begin{tabular}{ll}
\textbf{Contrastive Transformations}
& 
\textbf{Parameter}\\
\hline
Random Crop \& Resize (percentage) &
0.4 - 0.8\\
Random Horizontal Flip (probability) &  0.5\\
Random Color Jittering (probability &0.8 \\ 
Random Gaussian Blur (kernel size) &  21 \\
Random Solarize (threshold) &  0.5 \\
Random Rotation (degrees) &  $\pm5$ \\
Random Translation (percentage) & $\pm$0.05 \\
Input Time Difference &  1-12 Months \\
Normalization Mean \& std &  (0.202, 0.113) \\
\hline
OCT Scanner & Cirrus OCT\\
Resolution & $6\times6\times2$ mm$^3$\\
\hline
Projector MLP dimensions & 4096-4096-4096\\
Predictor $\rho_\psi$ MLP dimensions& 2049-2048\\
\hline
GPU Used & Nvidia A100 80GB\\
\end{tabular}
\end{center}
\end{table}
\subsection{Non-Contrastive VicReg Loss}\label{vic}
VICReg loss with \textit{invariance}, \textit{variance}, and \textit{covariance} terms, is as follows:
\begin{align} \label{eq:invariance}
     \ell_{\mathrm{contr}}(Z_t, Z_{t+\Delta t}) &= \lambda_S \cdot S(Z_t, Z_{t+\Delta t}) + \lambda_V \cdot (V(Z_t) + V(Z_{t+\Delta t})) \\ &+ \lambda_C \cdot (C(Z_t) + C(Z_{t+\Delta t}))\nonumber \\
     S(Z, Z^\prime) &= \frac{1}{n} \sum_i \|z_{i} - z^\prime_{i}\|_2^2 \nonumber 
     \quad V(Z) = \frac{1}{d} \sum_{j=1}^{d} max(0, 1 - \mathrm{std}(z^{j}, \epsilon))\nonumber \\
     C(Z) &= \frac{1}{d} \sum_{i \ne j} [Cov(Z)]_{i,j}^2, \nonumber\\ \textrm{where} \ \ \  Cov(Z) &= \frac{1}{n - 1} \sum_{i=1}^{n} (z_{i} - \bar{z})(z_{i} - \bar{z})^{T} \nonumber .
\end{align}
 $\lambda_S$, $\lambda_V$, and $\lambda_C$ are set to 15, 25, 5 to bring their magnitude in the same range. 
\subsection{Derivation of Displacement Map Regularization Term}
{The pairwise RankNet~\cite{burges2005rank} loss is based on cross-entropy loss for calculating the score-based ranking a pair of elements. Let $f$ be a scoring function for the items $x_i$ and $x_j$, such that their respective scores are $s_i = f(x_i)$ and $s_j = f(x_j)$. Concordantly the score difference is defined as $s_{ij}=s_i-s_j$. Then, the probability of ranking $x_i$ greater than $x_j$ is defined using the logistic function: 
\begin{equation}
    \label{eq:logprob}
   \mathcal{P}_{ij} = \frac{e^{s_{ij}}}{1 + e^{s_{ij}}} =  \frac{1}{1 + e^{-s_{ij}}}
\end{equation}
When ranking the items, there are 3 values for ground truth $Y$; 1 when  $x_i$ has higher rank than $x_j$, 0 when the relation is reversed, and $\frac{1}{2}$ when both items have the same rank. Accordingly, cross-entropy loss for correctly ranking the items is:
\begin{equation}
    \label{eq:ce}
   \mathcal{L}_{ce} = -Y \cdot \log(\mathcal{P}_{ij}) -(1-Y)\cdot \log(1-\mathcal{P}_{ij})
\end{equation}
In TC, $s_i$ becomes $r_{t+\Delta t}$ and $s_j$ becomes $r_t$, hence $s_{ij}$ is the DM. For the cross-entropy loss, we implemented $s_{ij}$ as the norm of the DM. Also the ranking class is always $1$ because the time difference is kept positive, thus the Eq.~\ref{eq:ce} becomes: 
\begin{align}     
    \label{eq:ce_reg}
   \mathcal{L}_{reg} &= -1 \cdot \log(\mathcal{P}_{ij}) -(0)\cdot \log(1-\mathcal{P}_{ij}) =  -\log \left( \frac{1}{1 + e^{-s_{ij}}}\right) \\
   &= \log( 1 + e^{-s_{ij}})= \log \left(1 + \exp(-||  \rho_\psi(r_t,\Delta t) ||_2) \right) \nonumber
\end{align}
\begin{table}[h]
\caption{Computational Costs of Each Model}
\label{Tbl:costs}
\begin{center}
\begin{tabular}{lll}
\hline
Model& 
\textbf{Batch Update (Seconds)}& 
\textbf{\# Trainable Parameters}\\
\hline
VICReg& 9.3& 65.4M \\
ESSL &  16.3& 78.0M\\
AugSelf &14.2 &78.0M\\ 
EquiMod & 12.6 &  124.2M\\
TC&  10.4 & 82.2M\\
\hline
\end{tabular}
\end{center}
\end{table}

\begin{figure}[b]
\begin{center}
\includegraphics[width=0.65\linewidth]{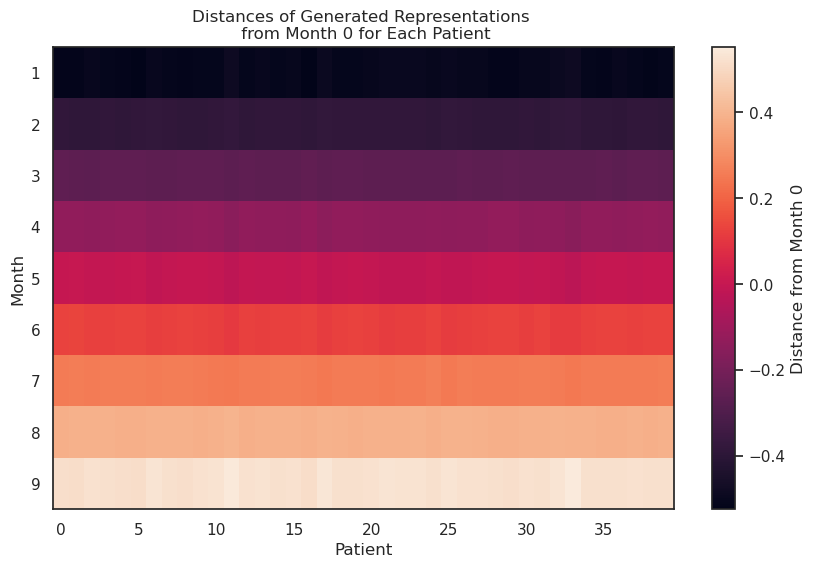}
\end{center}
\caption{Distance rankings between the original reprensentation $r_0$ and its propoageted prediction $r'_{t+ \Delta t}$ for 9 consecutive months for 40 patients.}
\label{fig:dist}
\end{figure}

\end{document}